\documentclass[11pt]{interflam}

\usepackage[numbers,sort&compress]{natbib}
\newcommand{\supercite}[1]{\textsuperscript{\cite{#1}}} 

\usepackage{tabularx}
\usepackage{ragged2e}
\usepackage{booktabs}
\usepackage[labelformat=simple]{subcaption}

\usepackage{textcomp}
\usepackage{gensymb}
\usepackage{siunitx}
\usepackage{xspace}
\usepackage{todonotes}
\usepackage[fleqn]{mathtools}
\usepackage{cuted}
\usepackage{placeins}
\usepackage[acronym]{glossaries}
\usepackage[T1]{fontenc}
\usepackage{amsmath}

\makeatletter
\renewcommand\AB@affilsepx{, \protect\Affilfont}
\makeatother

\newcommand{\ecoclient}{ecoClient\xspace}
\newcommand{\nosim}{noSim\xspace}
\newcommand{\optsim}{optSim\xspace}
\newcommand{\carbondioxide}{CO$_2$\xspace}
\newcommand{\fortran}{Fortran\xspace}

\newcommand{\unet}{U-net\xspace}
\newcommand{\unets}{U-nets\xspace}

\newcommand{\brac}[1]{\left({#1}\right)}

\usepackage[acronym]{glossaries-extra}
\makenoidxglossaries

\newacronym{cfd}{CFD}{Computational Fluid Dynamics}
\newacronym{fair}{FAIR}{Findable, Accessible, Interoperable, and Reusable}
\newacronym{fno}{FNO}{Fourier Neural Operator}
\newacronym{pde}{PDE}{Partial Differential Equation}
\newacronym{fds}{FDS}{Fire Dynamics Simulator}
\newacronym{hpc}{HPC}{High Performance Computing}
\newacronym{ml}{ML}{Machine Learning}
\newacronym{hrr}{HRR}{Heat Release Rate}
\newacronym{vae}{VAE}{Variational Autoencoder}

\title{{Leveraging AI modelling for FDS with Simvue: monitor and optimise for more sustainable simulations}}

\author[a]{\underline{James Panayis}}

\author[a]{Matt Field}

\author[a,b]{Vignesh Gopakumar}

\author[a]{Andrew Lahiff}

\author[a]{Kristian Zar\k{e}bski}

\author[a]{\\Aby Abraham}

\author[c]{Jonathan L. Hodges}

\affil[a]{UK Atomic Energy Authority, UK} \affil[b]{UCL Centre for AI, UK}
\affil[c]{Jensen Hughes, USA}
\date{}

\begin{document}

\maketitle

\section{Abstract}\label{abstract}

There is high demand on fire simulations, in both scale and quantity. We present a multi-pronged approach to improving the time and energy required to meet these demands. We show the ability of a custom machine learning surrogate model to predict the dynamics of heat propagation orders of magnitude faster than state-of-the-art \gls{cfd} software for this application. We also demonstrate how a guided optimisation procedure can decrease the number of simulations required to meet an objective; using lightweight models to decide which simulations to run, we see a tenfold reduction when locating the most dangerous location for a fire to occur within a building based on the impact of smoke on visibility. Finally we present a framework and product, Simvue, through which we access these tools along with a host of automatic organisational and tracking features which enables future reuse of data and more savings through better management of simulations and combating redundancy.

\section{Introduction}\label{intro}

A key objective in fire safety design is to ensure that occupants of a building are able to vacate safely during an emergency. Removing smoke through exhaust systems which activate after detection of a fire is one approach to improving the safety of occupants during egress. These systems often consist of mechanical fans positioned near to the ceiling creating a current of air drawn from floor-based inlets. 

The placement of such systems and the air flow rates required to maintain tenable conditions varies depending on application. These systems are designed by following available guidance, with sizing and placements for the final design often being refined and tested through \gls{cfd} fire modelling. This modelling allows an engineer to assess the flow of heat, smoke, and products of combustion throughout an enclosure within a fire scenario. Such simulations take a significant time for completion, typically 500-1000 core hours (runtime multiplied by number of cores), with larger, more complex domains such as car parks and subways having considerably larger requirements.

Parametric studies varying ventilation locations and flows are often run in parallel to meet project deadlines, consuming a high volume of resources. To reduce this, there are computational tools which are able to accelerate model execution by intelligently selecting parameters for simulation, and monitor on-going simulations to prematurely abort those which have failed conditions of success. One such tool is Simvue, which further addresses data management and provenance concerns, tracking campaigns and runs to aid reproducibility and facilitate reuse and verification of results and processes. In a world with ever-more data, planning and structuring of data systems, workflows, and metadata is an increasingly pressing concern to ensure that data is \gls{fair} \supercite{Wilkinson2016} — not just for humans, but also for machines and scientific \gls{ml}.

Physics-informed, data-driven surrogate models can efficiently approximate the behaviour of complex dynamical systems~\supercite{PIML}. Neural operators (a class of neural networks that learn mappings between function spaces) have demonstrated remarkable capabilities in modelling \glspl{pde}, making larger ensembles feasible and thus facilitating investigations into higher-scale and more extreme scenarios ~\supercite{Brunton2024, neural_operator}. The \gls{fno}, an efficient variant of neural operators, has proven valuable across diverse modelling applications, from weather forecasting to nuclear fusion, demonstrating quicker training with reduced data requirements than alternatives.~\supercite{pathak2022fourcastnet, Gopakumar_2024} These surrogate models have achieved remarkable computational efficiency, reducing simulation time by up to six orders of magnitude while enabling intelligent design space exploration that requires ten times fewer simulations.~\supercite{Guilhoto2024} Here, we investigate how \gls{ml}-driven \gls{pde} modelling can serve as an effective surrogate and optimisation toolkit for the fire modelling community.

The next section provides an overview of our case study. The three subsequent sections detail our key contributions. First, we introduce the headline capabilities of Simvue. Second, we present an optimisation example that significantly reduces the number of simulations required to identify the most hazardous fire location within a building. Third, we demonstrate that a surrogate model can accurately replicate the multidimensional temperature dynamics of original simulations while operating several orders of magnitude faster.

\section{Scenario}\label{scenario}

The investigations outlined within this paper all concern a specific building setup based on a round robin case study from the 2014 Society of Fire Protection Engineers (SFPE) Performance Based Design (PBD) Conference.~\supercite{sfpe2014pbd} As documented in the design brief, the building is a corporate headquarters office which contains a total of 8 floor levels (L1 to L8). L1 to L3 are car parking and L4 to L8 are open plan
offices. The basic building foot print is \SI{100}{\metre}$\times$\SI{30}{\metre} yet the floor plates exhibit heterogeneous and non-symmetrical layouts. The floor-to-floor height is \SI{3}{\metre} for the parking levels and \SI{4}{\metre} for
each of the office levels. The models in this work consider fires in the open plan offices on L4 to L8 (see \autoref{fig:unet_fds}).

\begin{figure}[h!]
    \centering
    \begin{minipage}{0.465\textwidth}
        \centering
        \includegraphics[width=\linewidth]{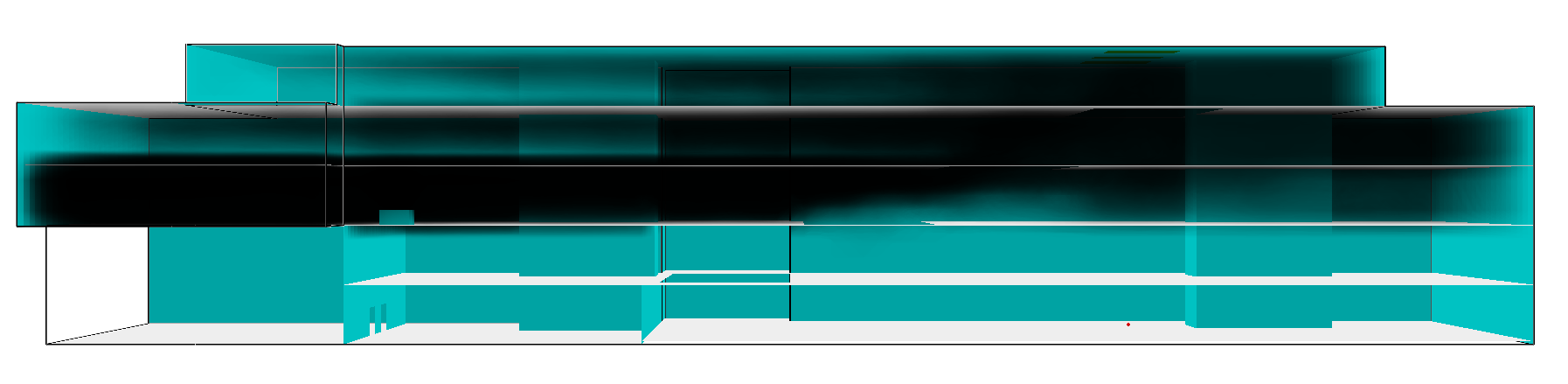}
    \end{minipage}
    \hfill
    \begin{minipage}{0.465\textwidth}
        \centering
        \includegraphics[width=\linewidth]{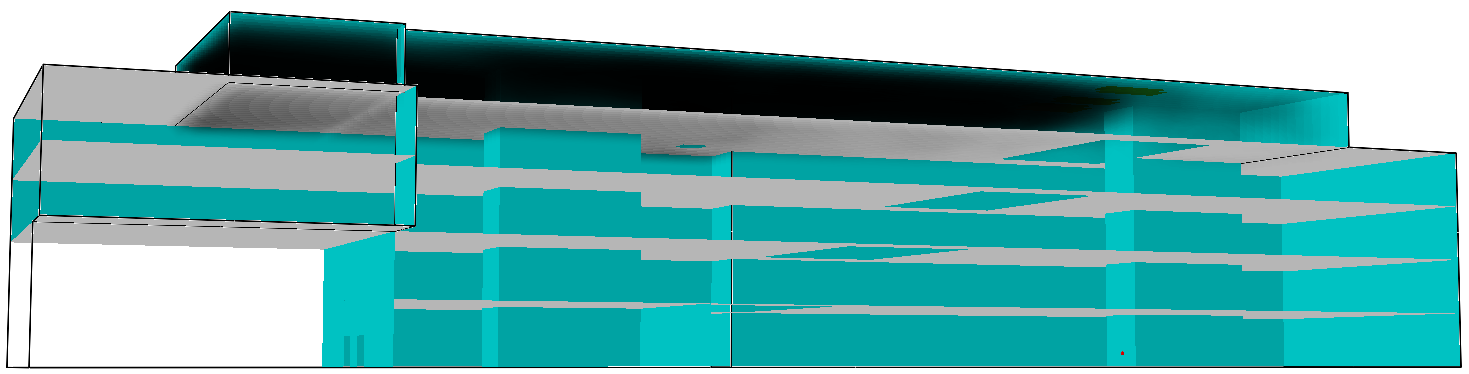}
    \end{minipage}
    \begin{minipage}{0.465\textwidth}
        \centering
        \includegraphics[width=\linewidth]{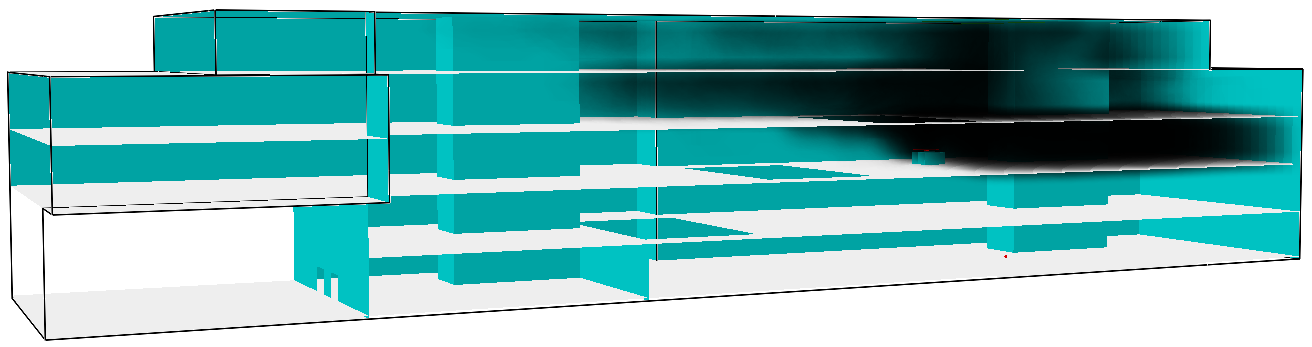}
    \end{minipage}
    \hfill
    \begin{minipage}{0.465\textwidth}
        \centering
        \includegraphics[width=\linewidth]{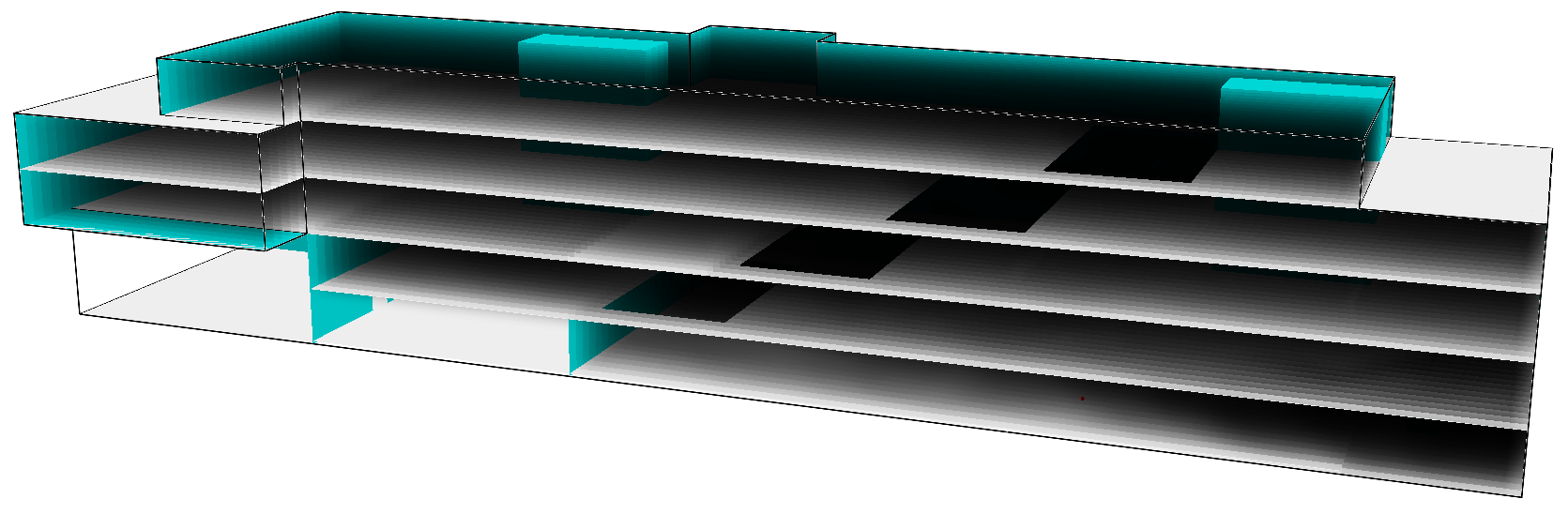}
    \end{minipage}
    
    \caption{Examples of various fire locations at different times simulated with \gls{fds}.}
    \label{fig:vis}
\end{figure}

There is a cascading sequence of floor openings which form communicating spaces between adjacent floors in a diagonal vertical path (see \autoref{fig:vis}) which facilitates the rapid movement of smoke between the levels. There are also two large vertical pillars that penetrate vertically through all the floors, further complicating the dynamics.

After a period of \SI{120}{\second} from the start of the fire, $3$ vents in the ceiling open and begin to mechanically extract air. These vents are located roughly above the communication area between the top two floors, and are intended to draw up air through the diagonal channel to clear smoke from the floors below. When the vents open, doors on the ground floor also open (near to the communicating region between the bottom two floors), allowing air to enter into the building, replacing that which is extracted.

The heat release rate is modelled as \SI{150}{\second} t-square growth to a peak fire output of \SI{1}{\mega\watt}. The fire is located at floor level, covering \SI{2}{\metre}$\times$\SI{2}{\metre} of floor area, and is placed at various positions throughout the different simulations, at least \SI{1}{\metre} away from the walls and pillars. The positions are determined according to a distribution that is uniform over the available floor area (except in specified cases), meaning that those floors with a larger area are overall more likely to contain a fire than those with a smaller area.

Each run simulates \SI{30}{minutes}, using a relatively course, uniform grid where the cells are cubes of side length \SI{1}{\metre}. While industry practice varies significantly in terms of grid resolution and simulation duration, we adopt this configuration throughout to ensure consistency and facilitate comparative analysis of results, as the techniques evaluated here scale effectively. We extract visibility distance data from \gls{fds} at every time-step. These data are time-averaged into $120$, \SI{15}{second} ranges, removing small-scale fluctuations which are not important to the models. We then specifically take the visibility distance data at approximately eye-level, \SI{2}{\metre} above each floor. This results in 5 slices through the dataset, each one the shape of the floor-plan of one floor, for each of the $120$ time values.


\section{Simvue}\label{sec2}

Computational speed-ups using \gls{ml} models often involve running simulation software on \gls{hpc} clusters consisting of thousands of CPU cores, interfaced by command line tools. Many of these codes are built using Python, C++, or \fortran, the latter two frequently used in high computational cost applications. These systems generally lack user-friendly interfaces, requiring users to manually edit files and input parameters through terminal-based workflows. Typically, when simulations are performed across a range of experimental parameters, some of these runs work whereas others crash. The output and system files associated with the simulation are usually stuck within the system, and thus relatively inaccessible, especially in error cases. Hence interacting with these codes requires extensive in-depth expertise and familiarity with simulation management.

\begin{figure}[h!]
    \centering
    \includegraphics[width=\linewidth, keepaspectratio]{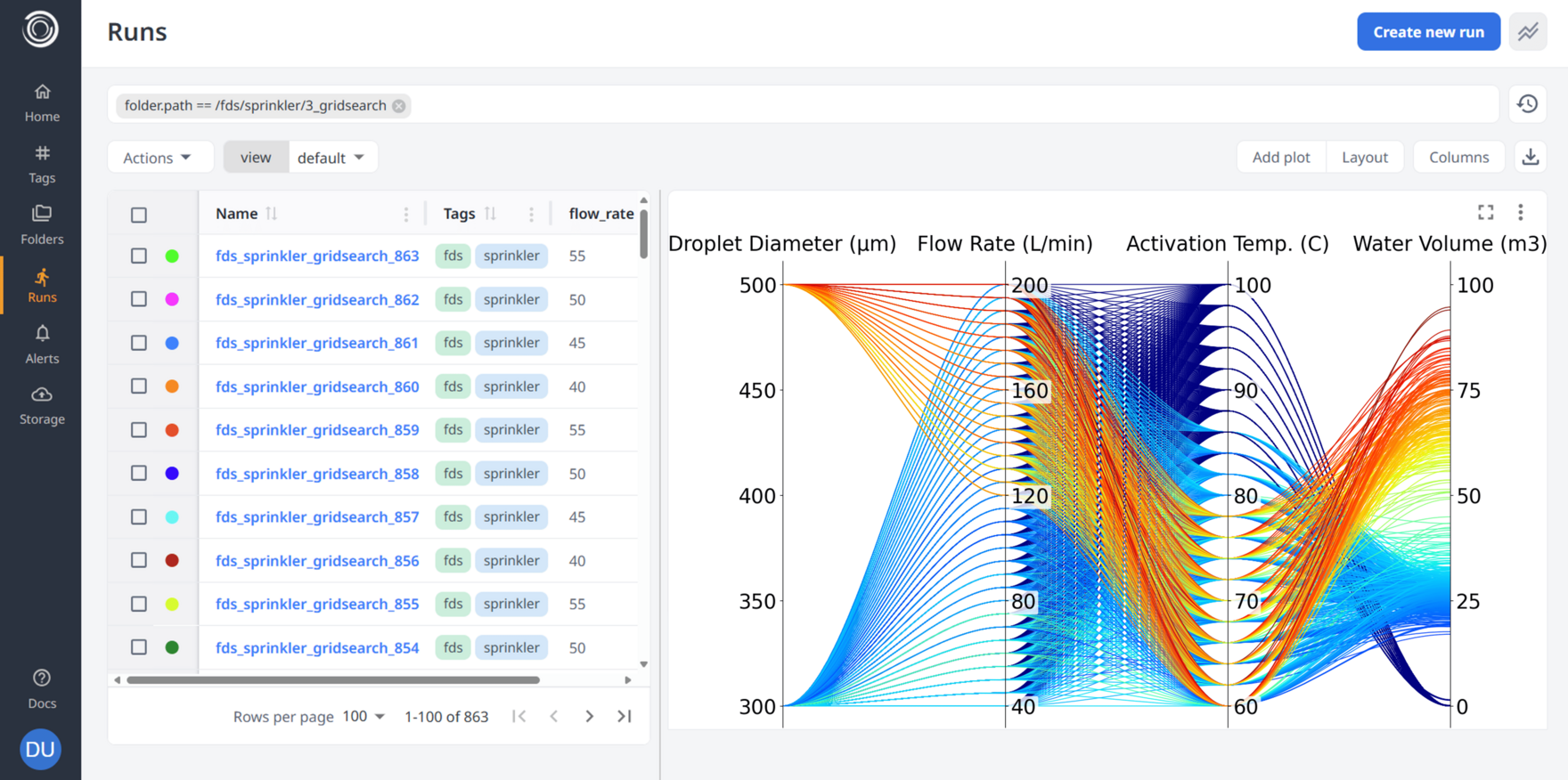}
    \caption{The Simvue dashboard, showing a collection of \gls{fds} simulations from a sprinkler design optimisation campaign. A parallel coordinates plot has been created using the metadata and metrics which are automatically tracked by Simvue.}
    \label{fig:dashboard}
\end{figure}

Simvue was developed to address these challenges; it is a framework that can monitor the progress and performance of simulations in real-time, automatically tag metadata within campaigns like parameter scans, and store the simulation inputs and outputs in an interactive dashboard supported by a dependency graph. An image of this dashboard displaying a number of \gls{fds} runs is shown in ~\autoref{fig:dashboard} - runs have been sorted into folders and tagged with appropriate metadata, which can then be displayed and filtered on within the summary dashboard. 

Simvue can also be used to develop quickly interactive plots to gain an understanding of the data, such as the parallel coordinates plot shown on the right of ~\autoref{fig:dashboard}. The example highlights the link between sprinkler design parameters and total water usage to extinguish a fire, with each line representing a single simulation. Simvue extends beyond the simulation space to create \gls{ml} models and craft optimal experiment design strategies that utilise an efficient trade-off between exploring unknown parts of the parameter space and exploiting parts of it known to be desired. This framework is applicable to both current and legacy applications. The platform is also language- and hardware-agnostic, and capable of scaling on any cloud based or bare-metal architecture.

\subsection{Data Lineage}
\begin{figure}[h!]
    \centering
    \includegraphics[width=\linewidth, keepaspectratio]{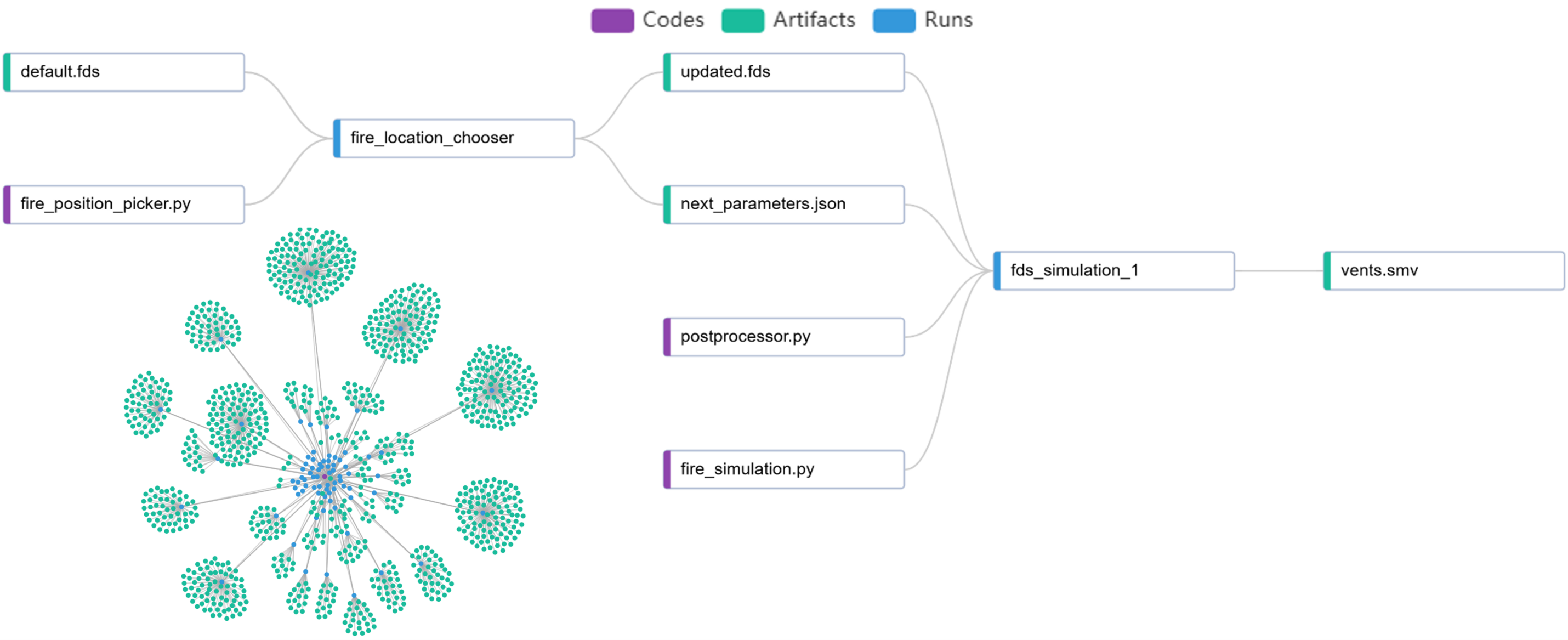}
    \caption{Diagrams showing data traceability with Simvue. The dependencies graph (bottom left) shows a series of runs which all depend on the same code file and each produce their own cloud of output artifacts. The lineage graph shows the files and runs required to generate a selected Smokeview file.}
    \label{fig:enter-label_a}
\end{figure}

Simvue’s data lineage feature enables full traceability of inputs, parameters, and outputs in simulations and data processing tasks, making it easy to understand how results were generated (see \autoref{fig:enter-label_a}). By automatically capturing metadata and supporting structured tagging of experiments, Simvue helps researchers organize data systematically and consistently. This streamlines the process of making data follow the \gls{fair} principles, ensuring that simulation results can be audited, reused, and shared across teams and projects with confidence.

\subsection{Real-time Monitoring}
\begin{figure}[h!]
    \centering
    \includegraphics[width=\linewidth, keepaspectratio]{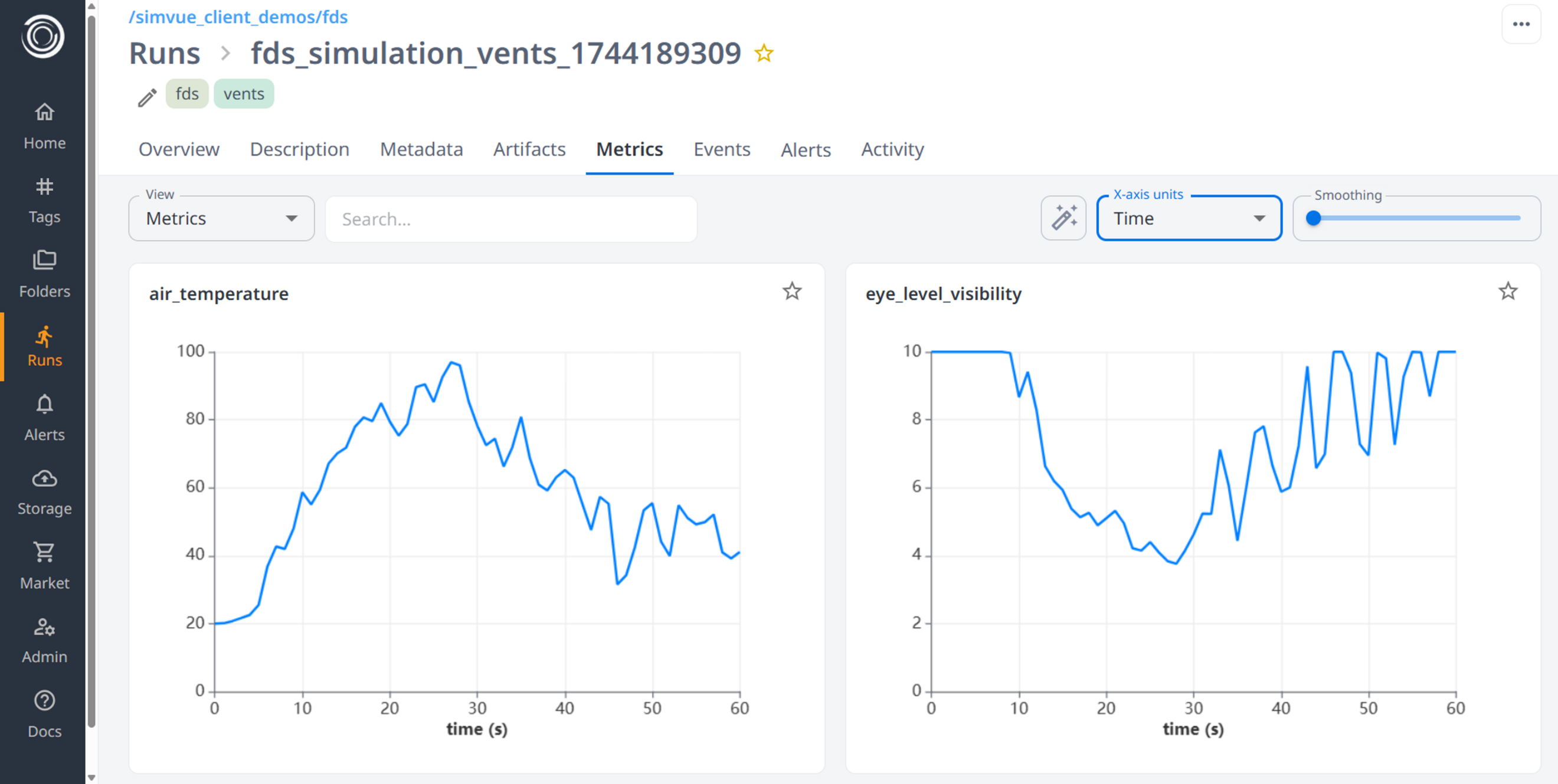}
    \caption{Real-time monitoring of a live \gls{fds} simulation.}
    \label{fig:enter-label_b}
\end{figure}

Simvue’s real-time monitoring feature (see \autoref{fig:enter-label_b}) provides live insights into the progress and status of fire safety simulations run using \gls{fds}, allowing users to track key metrics e.g., \gls{hrr}, performance indicators, and resource usage while simulations run.

\subsection{Alerting and Early Stopping}
\begin{figure}[h!]
    \centering
    \includegraphics[width=\linewidth, keepaspectratio]{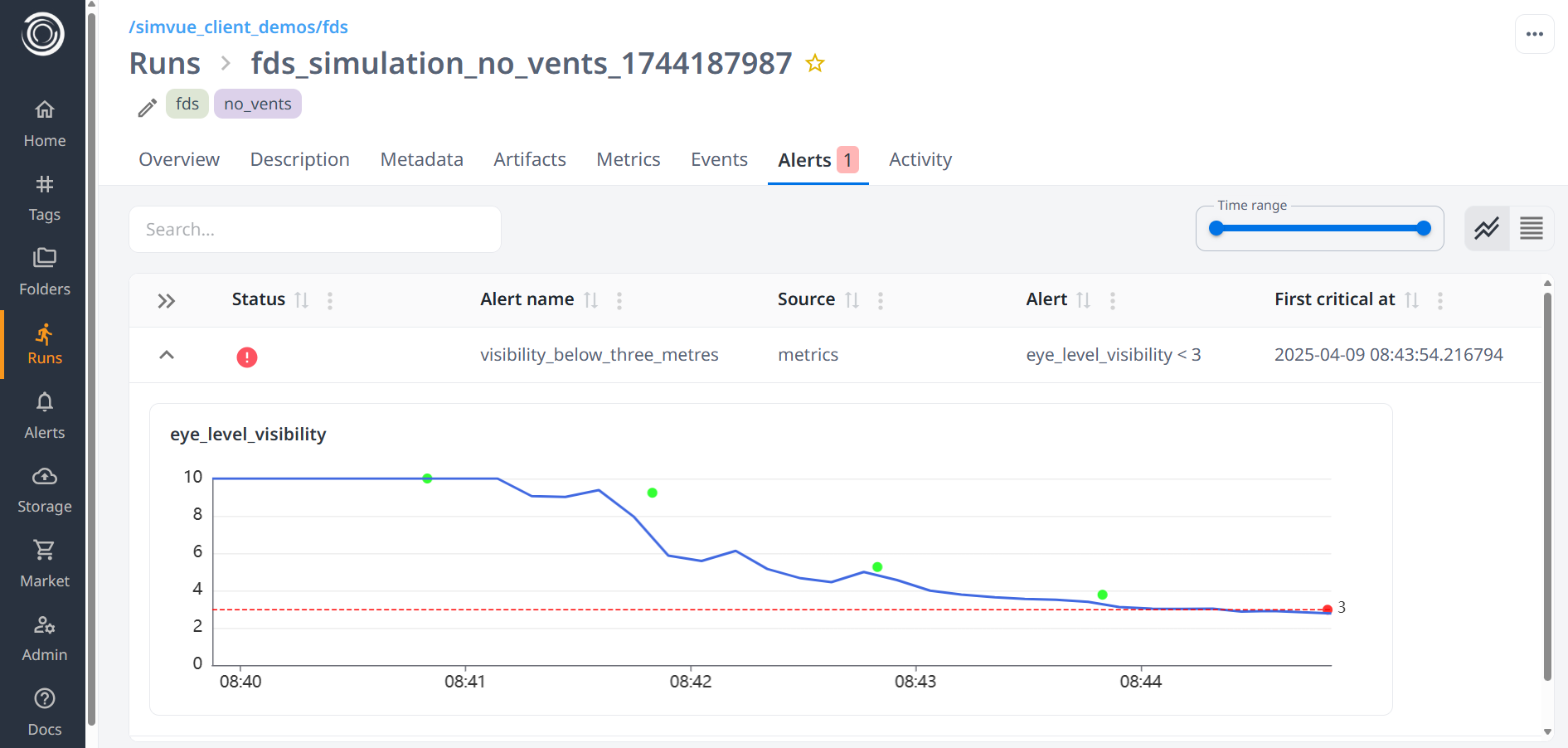}
    \caption{Alert notification and \gls{fds} runs terminated when visibility distance drops below \SI{3}{\metre}.}
    \label{fig:enter-label_c}
\end{figure}

By enabling automated alerting on thresholds or anomalies, Simvue helps researchers detect issues such as divergence, instability, or system failures early without waiting for completion. This proactive oversight is especially valuable for complex and long-running simulations, improving reliability, reducing wasted compute time, and supporting better decision-making during critical modelling phases. \autoref{fig:enter-label_c} shows an example of an early terminated \gls{fds} simulation when visibility drops below a threshold.
 
\subsection{Historic Data}
Simvue enables users to load any historic simulation run with full context including automatic capture of parameters, metadata, output, logs, and code version. This functionality allows researchers to reproduce, compare, or extend past analyses with confidence, there being no need to manually track inputs or configuration details. By making historical runs easily searchable and accessible, Simvue unlocks the potential for post-hoc analysis, model optimisation, and training of \gls{ml} models on curated high-quality data. This not only accelerates research workflows, facilitates robust and uncomplicated onboarding to Simvue.

\subsection{Collaboration and Sharing}
Collaborating on complex simulations such as \gls{fds} presents significant challenges due to the sheer volume of output files, intricate parameter sets, and the need for precise documentation to ensure reproducibility. Team members may struggle with tracking which inputs produced which outputs, managing multiple versions of configuration files, and sharing complete results in a usable form. This complexity can lead to confusion, loss of data, and inconsistent outcomes between collaborators.

Simvue addresses these challenges by capturing and organizing every simulation run with complete data lineage, including parameters, code versions, output files, and metadata. This allows users to share runs seamlessly, providing collaborators with full context and a structured, searchable record of each run. By supporting \gls{fair} data principles, Simvue ensures that \gls{fds} simulations are easy to revisit, compare, and build upon, fostering more reliable and efficient collaboration in high-stakes research environments.
 
\subsection{Distributed Computing}
Simvue incorporates seamless distributed computing to simplify workload execution, whether users are running a single job or thousands. It connects to public and private clouds, \gls{hpc} clusters, and other platforms, such as Slurm and Kubernetes. The framework enables users to define, execute, and monitor jobs across multiple resources, supporting workflows like using a local cluster by default and bursting onto a public cloud when capacity is exceeded.

\subsection{Integration}

To launch an \gls{fds} run without Simvue, the user will need to write an input file which includes details of their specific scenario, such as the structure of the mesh to simulate over, the location, fuel type, and intensity of the fire, and any mitigation strategies which they are analysing within the space. They will then run the \gls{fds} executable on this file to begin the simulation, outputting information to the console and creating dozens of results files. 

To streamline the tagging and storage of data produced by these simulations, as well as enable the tracking, monitoring, and optimization of key parameters in the simulation, a Python-based connector class has been created for \gls{fds}. Other such connectors also exist for a variety of software packages used in various domains and industries. The \gls{fds} connector takes the location of input file and launches the simulation (optionally across multiple CPU cores in parallel using MPI), automatically extracting key results from the files being produced in real time to give critical insights into the data with little overhead for the user. By default the connector will:
\begin{itemize}
\item Parse the input file (which is in the format of a \fortran\ Namelist) and upload its contents as metadata. This allows for filtering and analysis of simulation results based on input parameters.
\item Extract data from the DEVC CSV file~\supercite{mcgrattan2025} whenever a new result is written and uploads the values as metrics. This allows for the monitoring of devices defined by the user, such as the visibility at different points in the simulation space or the temperature recorded by a thermocouple, in real time as the simulation progresses.
\item Extract data from the \gls{hrr} CSV file~\supercite{mcgrattan2025} and uploads the values as metrics. This file records information such as the heat release rate of the fire, mass production rate of gas species and zonal background pressures during the simulation, allowing for tracking of these important parameters in real time.
\item Track information from the console output~\supercite{mcgrattan2025}, uploading information such as the version of \gls{fds} used as metadata and messages such as the current time step being simulated as events. This will also upload any errors reported by \gls{fds} on startup as events, allowing for the user to quickly identify failed runs and correct issues.
\item Monitor the DEVC and CTRL log file~\supercite{mcgrattan2025} to identify when any device or control functions changes its logical state (such as a timer activating or a vent opening), uploading information about the state of these as metadata and events.
\item Stores all input and output files as artifacts, which can be retrieved for further data analysis at a later time.
\end{itemize}

The connector can also be used to load historic \gls{fds} simulations into Simvue, if the results have been stored locally by the user. For this functionality, the user simply needs to provide the directory containing the results, and the connector will automatically extract the relevant data. This allows for experienced users of \gls{fds} to quickly benefit from Simvue's capabilities, without having to rerun any simulations.

\subsection{Optimisation}\label{opt}

Simvue provides a framework, \optsim, to allow multiple optimisation strategies to be easily utilised. Once a problem is formulated as a transformation from input parameters to evaluation parameters (to be minimised or maximised), a selected strategy can then be applied to search the input parameter space for optimal points. This can be a traditional method such as a grid-search, a more direct method like gradient descent, or an \gls{ml}-enabled method like Bayesian optimisation, which can balance exploration of the parameter space with exploitation of the areas known to produce favourable results.~\supercite{Jones1998} Users are able to interchange or mix-and-match these strategies, empowering even those with no prior experience using more `intelligent' methods access to leverage them in their industries and for their unique challenges.

In fire safety, such an optimisation campaign may be performed to determine optimal sprinkler parameters; for example, in \autoref{fig:dashboard} we see a campaign varying droplet size, flow rate, and activation energy to minimise total water usage while ensuring adequate fire management. Alternatively, physical parameters such as the size, position, or extraction rates of vents may be investigated, or more structural aspects such as the widths of doorways. The benefit of such a general platform is that it enables the same tools to be used for any adequately-specific task. In \autoref{optex} we demonstrate this versatility by instead using the framework to determine the most dangerous location for a fire to occur in a building.

A user may also wish to use an existing dataset to begin this search, to leverage past work more effectively. Similarly, it may be useful to build a model on this data, or data produced from another optimisation campaign. Once an \gls{fno}, for example, has been created from simulation results in this way, it can then be used as the transformation in a new optimisation campaign - one which generates results far faster than \gls{fds}. Thus this surrogate model allows a more efficient exploration of the parameter space, and \gls{fds} only need be re-employed once the surrogate suggests we have found a satisfactory solution. Bringing together all these tools in a user-friendly framework expands accessibility to a range of people and industries, letting them enjoy the economic benefits they can bring, without requiring expert experience.

\subsection{Numerical Modelling}\label{sec4}

\begin{figure}[htb]
    \centering
    \begin{subfigure}[b]{0.95\linewidth}
        \includegraphics[width=\linewidth]{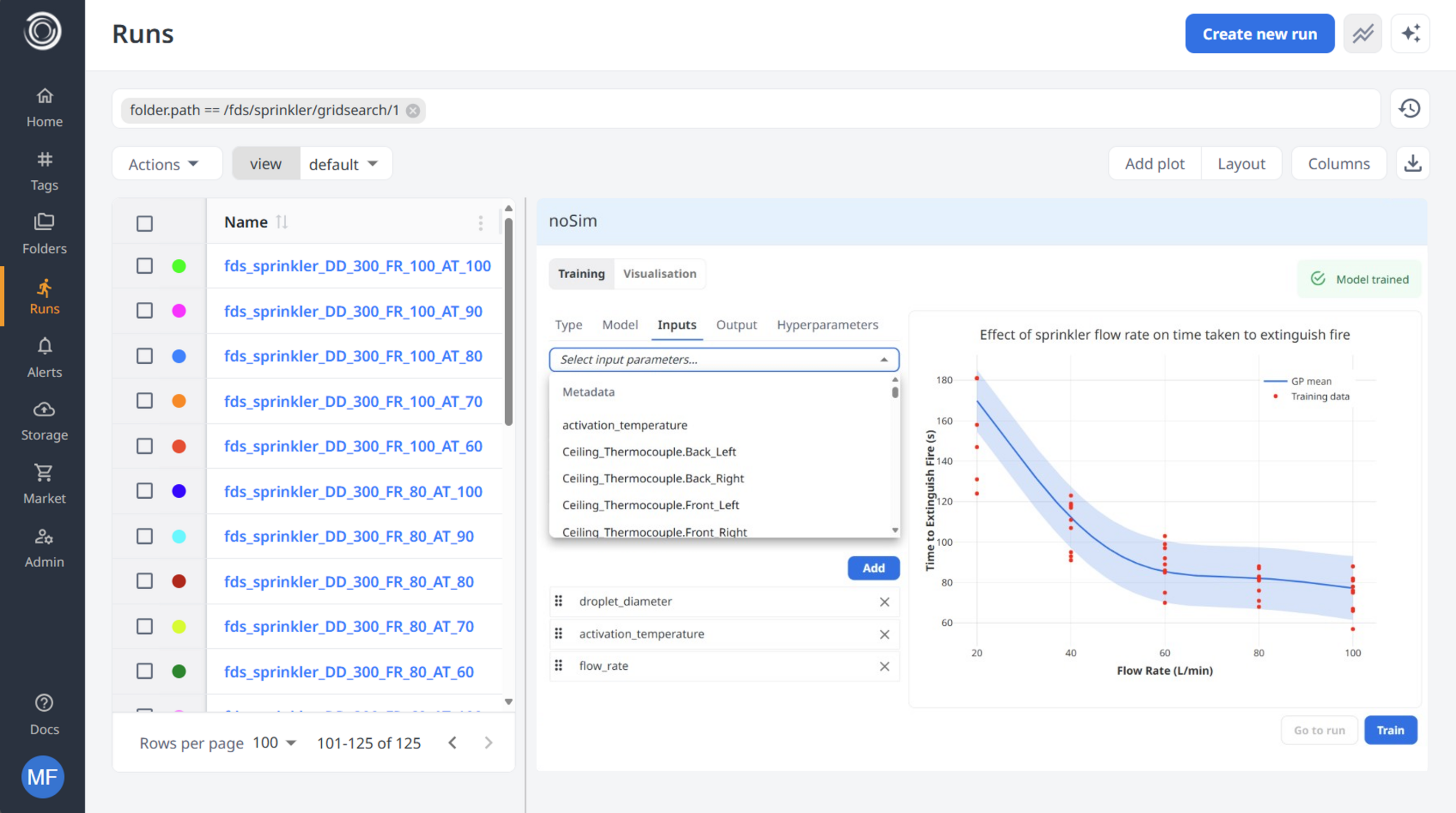}
        \caption{Setting up \nosim via the UI, and using the model to analyse Flow Rate vs Fire Extinguish Time}
    \end{subfigure}
    \begin{subfigure}[b]{0.45\linewidth}
        \includegraphics[width=\linewidth]{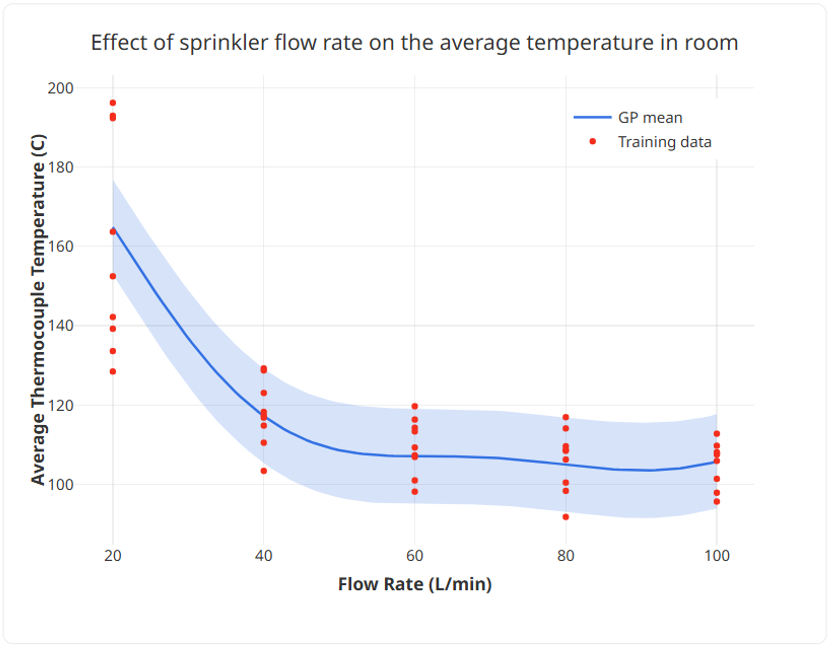}
        \caption{Flow Rate vs Average Temperature}
    \end{subfigure}
    \begin{subfigure}[b]{0.45\linewidth}
        \includegraphics[width=\linewidth]{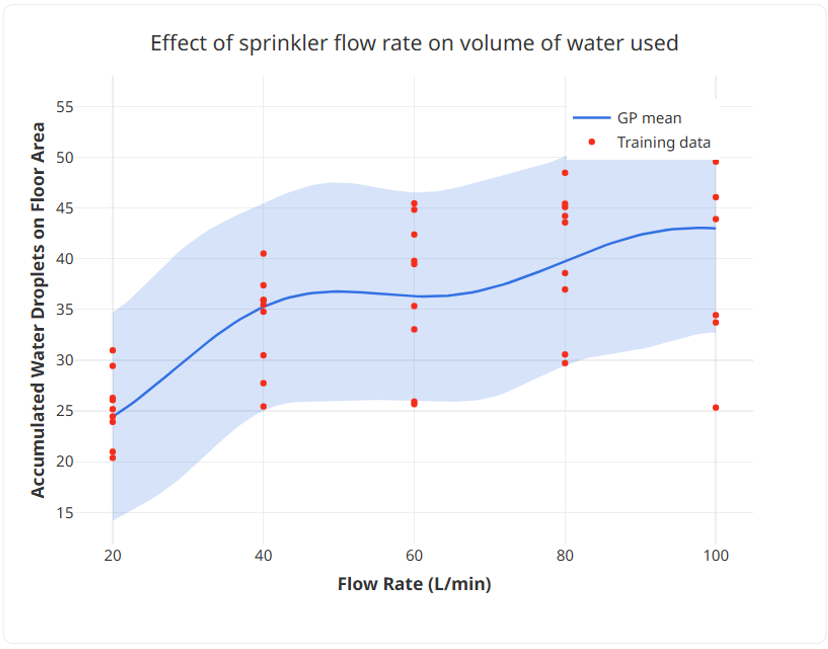}
        \caption{Flow Rate vs Water Usage}
    \end{subfigure}
    \caption{Using the \nosim\ functionality to train a Gaussian Process \gls{ml} model on data from a series of \gls{fds} sprinkler simulations. The user can gain quick insights into their data, and find the effect of varying different design parameters on the effectiveness of the sprinkler.}
    \label{fig:noSim}
\end{figure}

To help scientists and engineers gain insights into their data, Simvue also offers a feature called \nosim. This uses \gls{ml} models built into Simvue which are domain and simulation agnostic and can be accessed via the web UI. As the data from various simulations are stored in Simvue using \gls{fair} principles, the majority of the work needed to facilitate \gls{ml} modelling is already in place. Users can therefore specify the inputs and outputs which they care about, and a range of simulations to retrieve data from, finally selecting a model such as a Gaussian Process to be automatically trained. This model can then be used to identify key trends and form predictions for alternative simulations without needing to execute them, hence saving computational time and cost. 

As an example~\autoref{fig:noSim} shows the results of a simple simulation campaign which aims to optimise the design of a sprinkler by varying the activation temperature, droplet diameter and flow rate. Simulations are performed with five values for each of these parameters, and the results are then used to train a Gaussian Process. The model is then used to identify trends in the data, such as the effect of varying the flow rate of the sprinkler on the average temperature in the room, the volume of water used, and the time taken to extinguish the fire. A fire safety engineer can quickly interrogate this data, for example to identify the optimal flow rate based on their relative concern for these competing criteria.

\subsection{Measuring Carbon Footprint}

For complex \gls{fds} investigations, the resource utilisation of simulations becomes a significant metric in terms of environmental impact. Computing resources are heavy consumers of power and ultimately contribute to the emission of \carbondioxide due to the majority of energy is still produced from fossil fuels.

The \ecoclient extension for the Simvue interface provides the option to attach comparative estimates of the \carbondioxide emissions of runs; the calculation of these includes usage of carbon intensity metrics gathered from the Electricity Maps API.\supercite{electricitymaps2025}

\section{Model-Guided Optimisation}\label{optex}

The lower the visibility, the more difficult it is to evacuate a building and to send in firefighters to combat the fire. We therefore ask the pertinent question of where the worst possible position for a fire to occur is, such that planning can be coordinated around this worst-case scenario and mitigations can be enacted to address it. To answer this, we first define a scalar metric which converts the results of a simulation into a single number evaluating the `badness'; this is a partially subjective exercise, but on balance we opt to use

\begin{equation}
\sum_{i=0}^{119}\brac{ \frac{\text{count}\brac{\text{vis}_i<10}}{\text{count}\brac{\text{vis}_i}} + \alpha\max\brac{1-\frac{i}{80},0}\text{mean}\brac{\max\brac{20-\text{vis}_i, 0}} }
\end{equation}

The first term here measures the fraction of the area for which the visibility distance is less than \SI{10}{\metre}, accounting for the fact that we would like to keep that fraction as low as possible to maintain tenability as long as possible. For simplicity, this term is weighted equally across all $120$ time values (i.e. the whole \SI{30}{\minute} simulation); however, a non-uniform weighting could be used based on the objectives of the analysis, such as maintaining tenability through the required safe egress time (RSET). The second term measures how far below \SI{20}{\metre} the visibility has dropped on average, accounting for the additional difficulty posed to individuals attempting to evacuate. This is weighted down linearly with time from the start of the fire until \SI{20}{\minute} has passed, as the problems caused by such reduced visibility lessen as fewer people remain in the building. The second factor is further scaled by a factor, $\alpha$, which we set to $10$, to provide a reasonable balance between the two factors.

We first run $10$ simulations as described in \autoref{scenario}, with the fire locations chosen quasirandomly (via Sobol sampling quantised in the vertical direction). We then use \optsim~(\autoref{opt}) to decide which fire locations to simulate with \gls{fds} next. These locations are automatically chosen such that as much information as possible can be gleaned from the time and computation spent on the simulation. The process repeats, with \optsim discovering ever-worse fire locations. In total, $10$ extra simulations are performed, and the worst fire location was identified to be \SI{2.0}{\metre} from the left wall and \SI{10.6617}{\metre} from the front wall on the third floor. To achieve a similarly optimal result using a grid search, the grid would have to slice each floor into squares no larger than \SI{10}{\metre}$\times$\SI{10}{\metre}, requiring over $200$ simulations.

\begin{figure}[htbp]
    \centering
    \begin{subfigure}[c]{0.465\textwidth}
        \centering
        \includegraphics[width=\textwidth]{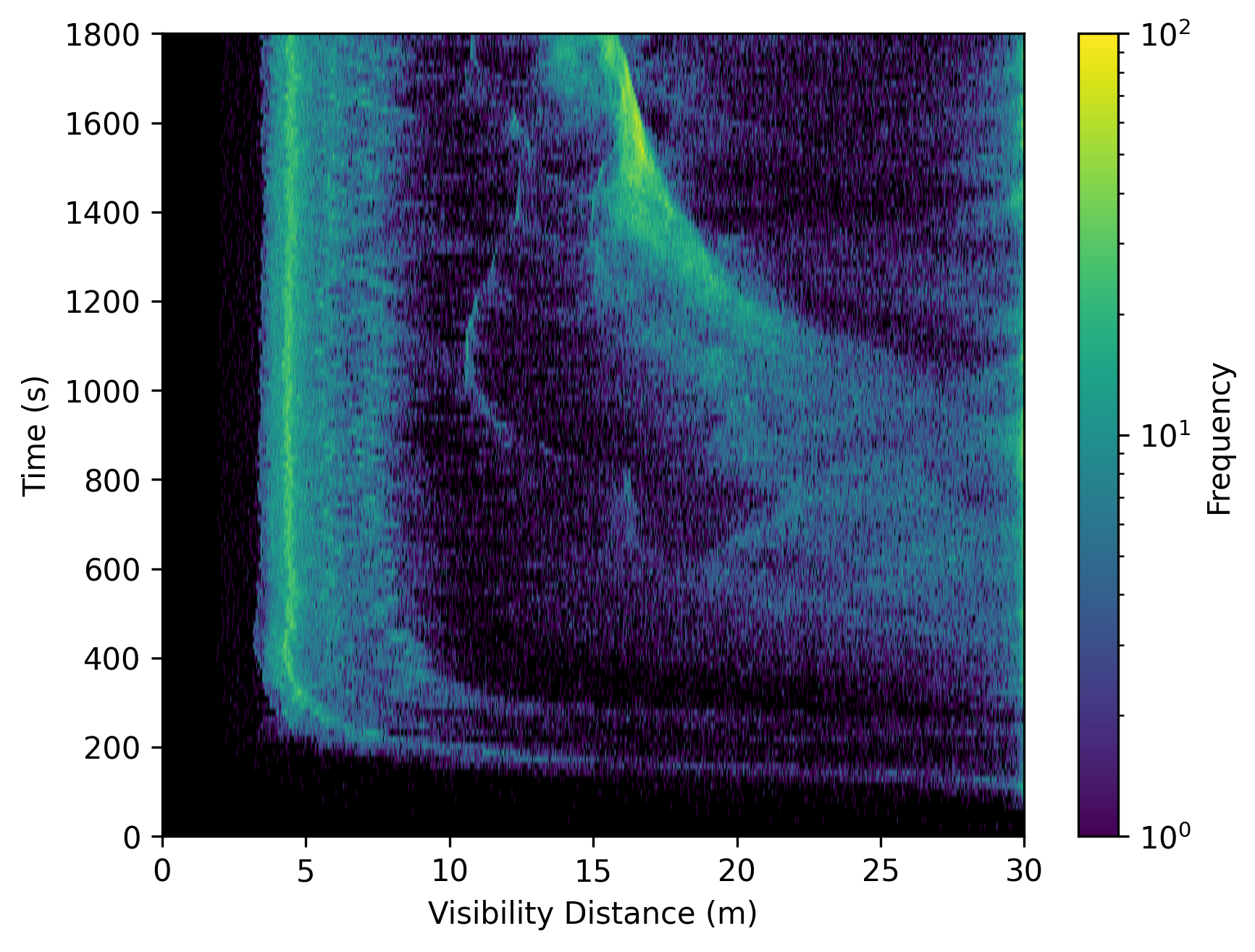}
        \caption{Evolution of the visibility through time. Each horizontal slice is a histogram of the visibility distances for a \SI{15}{\second} timeslice. Note that the maximum value that \gls{fds} is allowed to output in our simulations is \SI{30}{\metre}. This demonstrates that the visibility in a significant area (the left side of the third floor) quickly gets and then stays very low, while the visibility in another large area (the top floor) gradually and inconsistently decreases.}
        \label{fig: vis_evo}
    \end{subfigure}
    \hfill  
    \begin{subfigure}[c]{0.465\textwidth}
        \centering
        \begin{minipage}[c]{\textwidth}
            \centering
            \begin{subfigure}[t]{\textwidth}
                \centering
                \includegraphics[width=\textwidth]{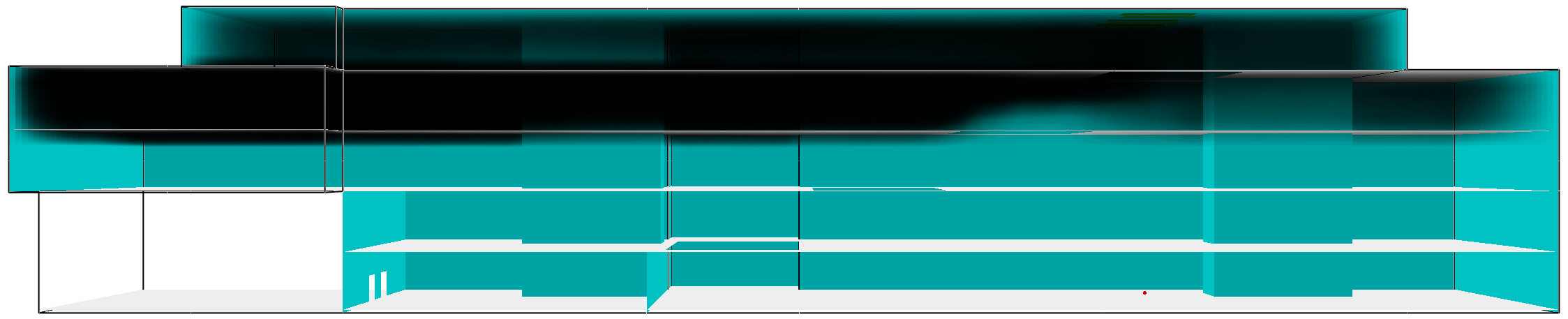}\par\vspace{0.5em}
                \includegraphics[width=\textwidth]{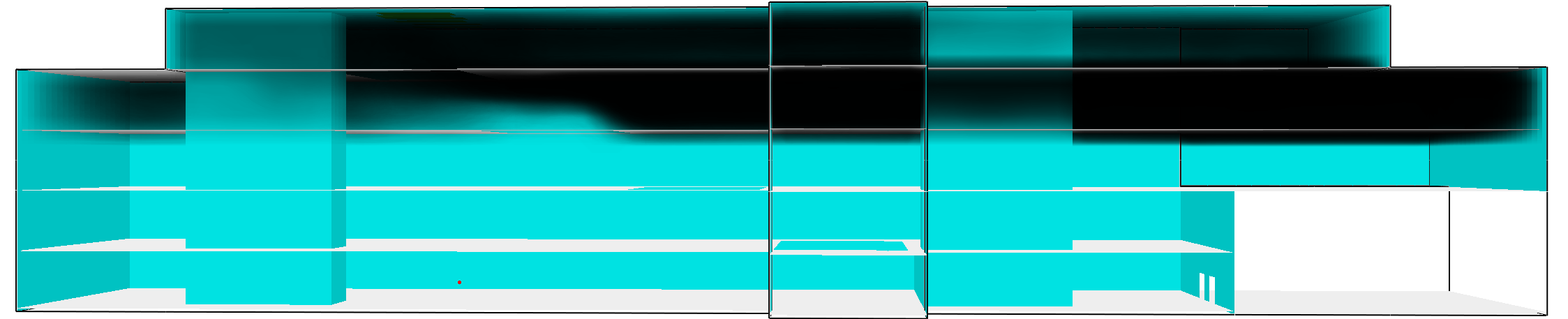}
                \caption{Front (top) and back (bottom) view of the distribution of smoke \SI{15}{\minute} into the simulation, demonstrating very high density on the third floor between the fire and the communicating regions, and significant though less dense smoke in the top floor above.}
                \label{fig: worst_smoke_building}
            \end{subfigure}
            \par\vspace{2em}
            \begin{subfigure}[t]{\textwidth}
                \centering
                \includegraphics[width=\textwidth]{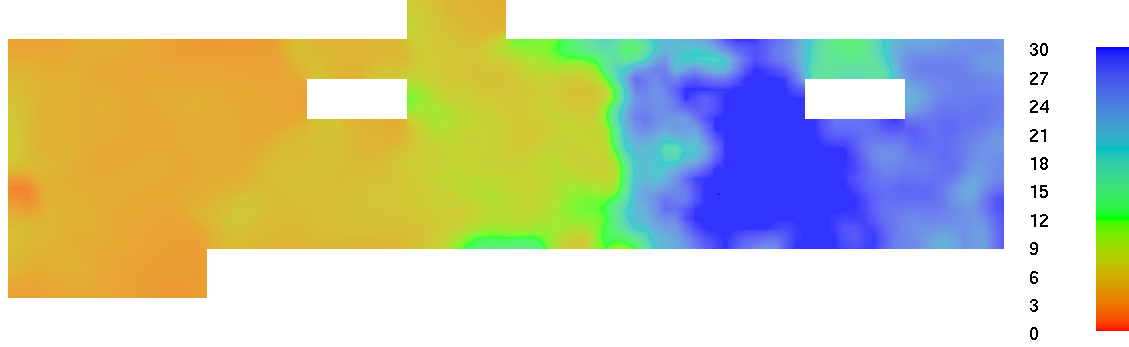}
                \caption{The visibility in metres after \SI{20}{\minute} at head-height above the third floor.}
                \label{fig: worst_smoke_single_floor}
            \end{subfigure}
        \end{minipage}
    \end{subfigure}
    \caption{Data from the simulation with the fire in the worst location - as close as possible to the leftmost wall of the third floor.}
    \label{fig:worst_fire}
\end{figure}

\section{Surrogate Modelling}
\label{sec:surrogat_modelling}
Surrogate modelling in fire safety simulation~\supercite{Marquant-2021} involves creating simplified, data-driven models that approximate the behaviour of computationally expensive simulations such as those run with \gls{fds}. These surrogate models—often built using \gls{ml} techniques like Gaussian processes, neural networks, or polynomial chaos expansions—can predict fire dynamics outcomes (e.g. temperature profiles, smoke spread, or evacuation times) rapidly, with minimal computational cost. This makes them valuable for tasks like sensitivity analysis, optimisation, and real-time decision support, where running full-scale \gls{cfd} models would be cumbersome or impractical. By training surrogate models on a well-structured set of simulation data, researchers can unlock faster scenario exploration, uncertainty quantification, and integration with control or risk assessment frameworks. Within Simvue we provide access to a range of modern neural PDE solvers such as \glspl{fno}\supercite{li2021fourierneuraloperatorparametric}, \unets\supercite{ronneberger2015unetconvolutionalnetworksbiomedical}, and conditional auto-encoders \supercite{kingma2022autoencodingvariationalbayes} for modelling the spatio-temporal evolution of smoke and the corresponding temperature. 

We train a custom \unet that learns to map the temperature distribution across all five floors (\SI{2}{\metre} above each floor) after \SI{20}{minutes} of the fire evolution, conditioned on the initial conditions that characterise the location and intensity of the fire. The neural network was trained using 200 simulations of \gls{fds} with varying fire locations. A \unet based neural operator was deployed for this problem to extract the highly localised behaviour often associated with fire dynamical modelling. \unets, structurally resembling a multi-grid method, allow for the extraction and distillation of the steep localised gradients characterising the smoke evolution.~\supercite{Hodges2019b, LE2021}

\begin{figure}[htb]
    \centering
    \begin{subfigure}[b]{0.85\textwidth}
        \centering
        \includegraphics[width=\textwidth]{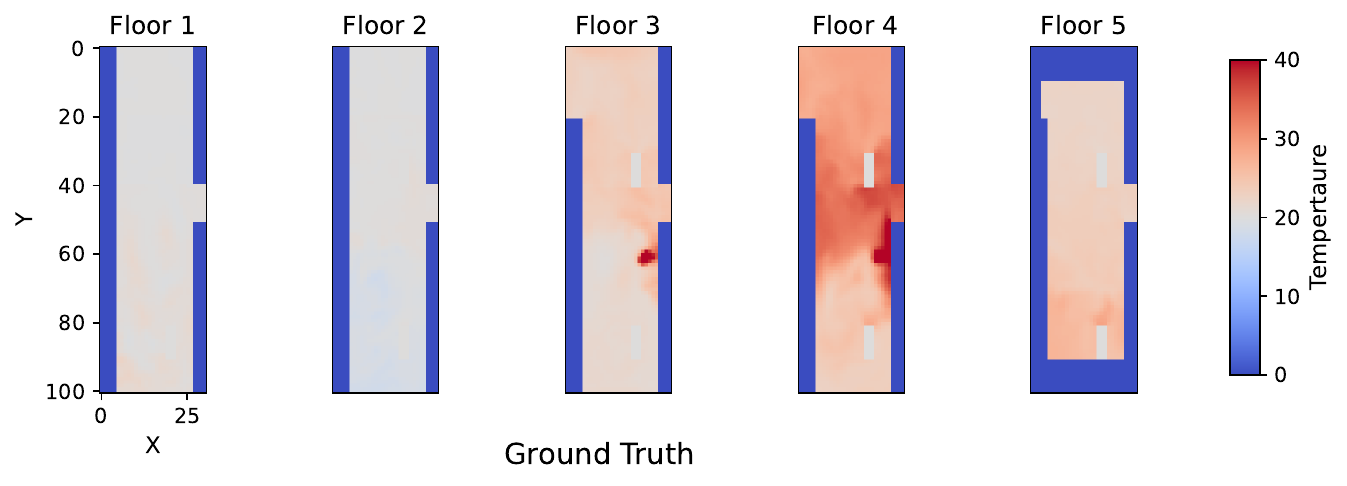}
        \caption{FDS simulation}
        \label{fig: FDS_sim}
    \end{subfigure}
    \\
    \begin{subfigure}[b]{0.85\textwidth}
        \centering
        \includegraphics[width=\textwidth]{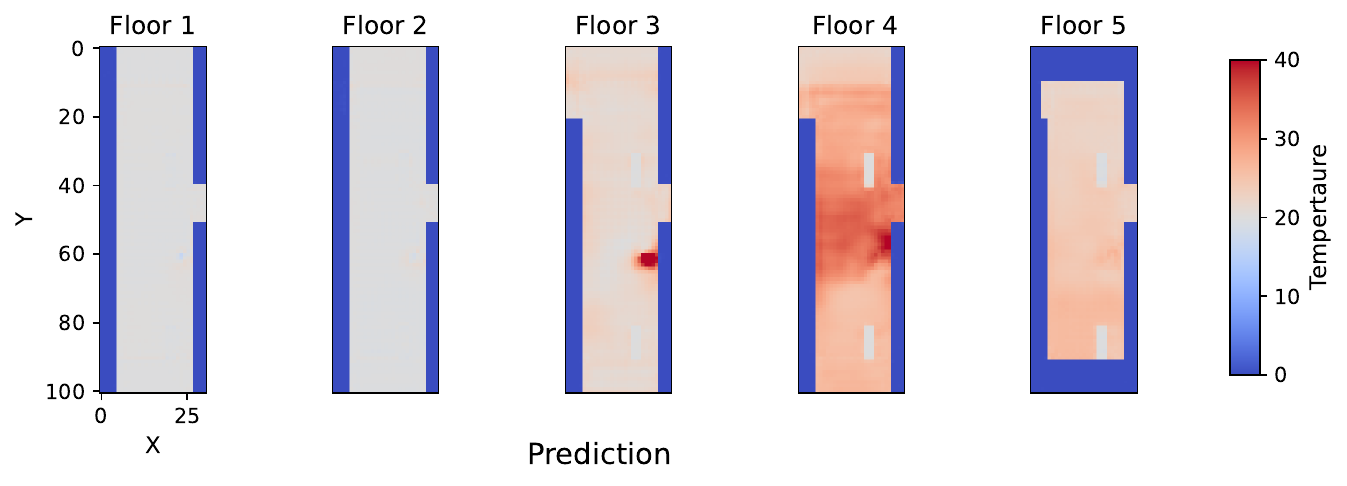}
        \caption{\unet prediction}
        \label{fig: Unet_pred}
    \end{subfigure}
    \caption{Comparing an \gls{fds} simulation with the \unet prediction of temperature distribution at the 20th minute across all 5 floors. The \unet is capable of modelling the larger features and capturing the smoke evolution at instantaneous speeds with considerable accuracy. }
    \label{fig:unet_fds}
\end{figure}

In~\autoref{fig:unet_fds} we compare the temperature distributions obtained by simulations with \gls{fds} and those obtained by predictions with our \unet. The \unet trained on $200$ simulations of \gls{fds} is validated over 50 simulations, yielding a mean squared error of \SI{2.815e-04}{} in the normalised space. The \unet is trained within \SI{5}{\minute} on an Nvidia A100 GPU, with the data extraction, processing, and training pipelines seamlessly set up by Simvue as given above. Once trained, the model can be evaluated almost instantaneously, yielding real-time approximations of the temperature, offering a speed increase of $\sim$4 orders of magnitude. 



\section{Conclusions}\label{sec5}

Improving the cost-effectiveness and sustainability of carrying out complex, long running \gls{fds} simulations is a growing concern within the fire safety community. To address this we demonstrate the advantage of modern surrogate models to significantly speed up the overall compute time. In addition we show the benefits of using Simvue to stop unnecessary and redundant long running simulations by using real-time tracking and monitoring. We also demonstrate how the framework ensures that all data generated follows the \gls{fair} principles, streamlining the storage, management, and retrieval of data to preserve lineages, foster transparency and collaboration, and enable the maximum value to be extracted from every simulation long into the future. As this is scaled up for numerous simulations in a campaign, this framework provides the necessary setup needed for \gls{ml} training and model development. Further, Simvue's \ecoclient functionality allows users to measure and keep track of the associated carbon footprint associate with each simulation.

Finally, we have shown the significant benefit of using Simvue to employ \gls{ml}-enhanced functionality. Using data to train surrogates allows a $10,000\times$ computational efficiency boost, replacing long-running simulations with rapid models. Similarly, using model-guided search methods allows us to optimise design parameters, offering a $10\times$ speedup over brute-forcing methods. The overall advantages, not just in time and cost savings but also reducing carbon footprint, make this a sustainable solution for rapid fire safety simulation. These improvements are so significant that they can not just transform existing procedures but also directly enable a new generation of techniques to be built atop them, driving further progress and evolution.



\section{References}\label{refs}

\bibliographystyle{ieeetr} 
\bibliography{references}

\section{Glossary}\label{gloss}
\vspace{-2.5em}
\renewcommand*{\acronymname}{}
\printglossaries


            

\end{document}